\documentclass[11pt, logo, onecolumn, copyright, colorlinks=true, allcolors=blue]{nvidiatechreport}

\usepackage{booktabs}
\usepackage{multirow}

\usepackage{graphicx}%
\usepackage{subcaption,ragged2e}
\usepackage[round]{natbib}
\usepackage{placeins}

\usepackage[normalem]{ulem}
\usepackage[utf8]{inputenc}%
\usepackage[T1]{fontenc}%
\usepackage{tikz}
\usetikzlibrary{arrows,calc}
\usepackage{pgf-pie}
\usepackage{enumitem}
\usetikzlibrary{positioning}
\usetikzlibrary{calc}
\usepackage{array}
\usepackage{wrapfig}
\usepackage{caption}%
\usepackage{listings}
\usepackage{adjustbox}
\usepackage{footnote}
\usepackage{amsmath}
\usepackage{amsfonts}
\usepackage{breakcites}
\usepackage[most]{tcolorbox}
\usepackage{tabularx}
\usepackage{xcolor}
\usepackage{float}
\usepackage{needspace}
\usepackage{makecell}
\usepackage{colortbl}
\usepackage{stfloats}
\setlist[itemize]{topsep=1pt,itemsep=1pt}
\setlist[enumerate]{topsep=1pt,itemsep=1pt}
\usepackage{xparse}
\usepackage{xspace}
\RenewDocumentCommand{\paragraph}{s o m}{%
  \par\medskip\noindent\textbf{#3}\quad\ignorespaces
}

\newcommand{\ourmodel}{Dory1.0-4B\xspace}
\newcommand{\ourbasemodel}{Dory1.0-4B Base\xspace}

\title{\ourmodel: A Looped Normalized-GPT }
\author{Maxim Fishman\textsuperscript{*}, Simeng Sun\textsuperscript{*}, Shantanu Acharya, Faisal Ladhak,  Brian Chmiel, Santiago Akle Serano, Ron Banner, Boris Ginsburg, Krishna C. Puvvada}
\renewcommand{\copyrightext}{\footerfont \textsuperscript{*}Equal contribution. \\ Correspondence to: \{mfishman,simengs,kpuvvada\}@nvidia.com \\ \textcopyright\, \the\year{} NVIDIA. All rights reserved.}
\date{}

\begin{document}

\begin{abstract}
\large \textbf{Abstract.}
\normalsize

We introduce \ourmodel, a 4-billion-parameter looped (latent-recurrent) language model with elastic depth: a single checkpoint serves different inference compute budgets. \ourmodel is built on a normalized Transformer (nGPT) backbone with interleaved sliding-window and full-attention layers, organized into an input block, a shared recurrent block, and an output block. At inference, the recurrent block can be applied a variable number of times. We pretrained \ourmodel on 6 trillion tokens. The first 4.5 trillion tokens were trained in NVFP4, and additional pretraining then extended the context length to 256K tokens. Next, we post-trained the model using Supervised Fine-Tuning (SFT) and Reinforcement Learning with Verifiable Rewards (RLVR), with a difficulty-aware loop curriculum that routes harder prompts to deeper loops. Increasing the loop count improves accuracy, with the largest gains on coding, agentic, and long-context benchmarks. With two or more loops, \ourmodel outperforms larger models such as Qwen3.5-9B, Gemma~4 12B, and Gemma~4 26B-A4B on average across 12 benchmarks. Looping also makes \ourmodel more token-efficient on agentic benchmarks: with four loops, it generates 1.3--2.5$\times$ fewer tokens than with one loop, as the extra computation is spent in latent space rather than on longer outputs. To reduce memory during looping, \ourmodel caches keys and values only from the last loop, keeping its KV-cache footprint equal to that of a non-looped model at any loop count while retaining most of the accuracy gains. We open-source the BF16, MXFP8, NVFP4, and GGUF quantized checkpoints.

\end{abstract}

\maketitle

\section{Introduction}
\label{sec:intro}

We present \ourmodel, the first open looped language model in the Nemotron Labs family. Scaling test-time compute has become a major driver of LLM capability, but this compute is mostly spent in token space: models generate longer reasoning traces, and latency, KV-cache footprint, and serving cost grow with every generated token. Looped (latent-recurrent) language models offer a complementary compute scaling axis~\citep{dehghani2019universal,saunshi2025reasoning,geiping2025scaling,zhu2025ouro}. By repeatedly applying a shared block of layers, a looped model spends more computation per token in latent space without adding parameters. \ourmodel brings this idea to a fully post-trained model built for reasoning and agentic tasks. We train \ourmodel with a mixture of loop counts~\citep{ladhak2026elr}, so that a single checkpoint can be served at any loop count $k$ and trade inference compute for accuracy per request, with no retraining and no model swap.

\begin{figure}[t]
    \centering
    \begin{subfigure}[t]{0.49\linewidth}
        \centering
        \includegraphics[width=\linewidth]{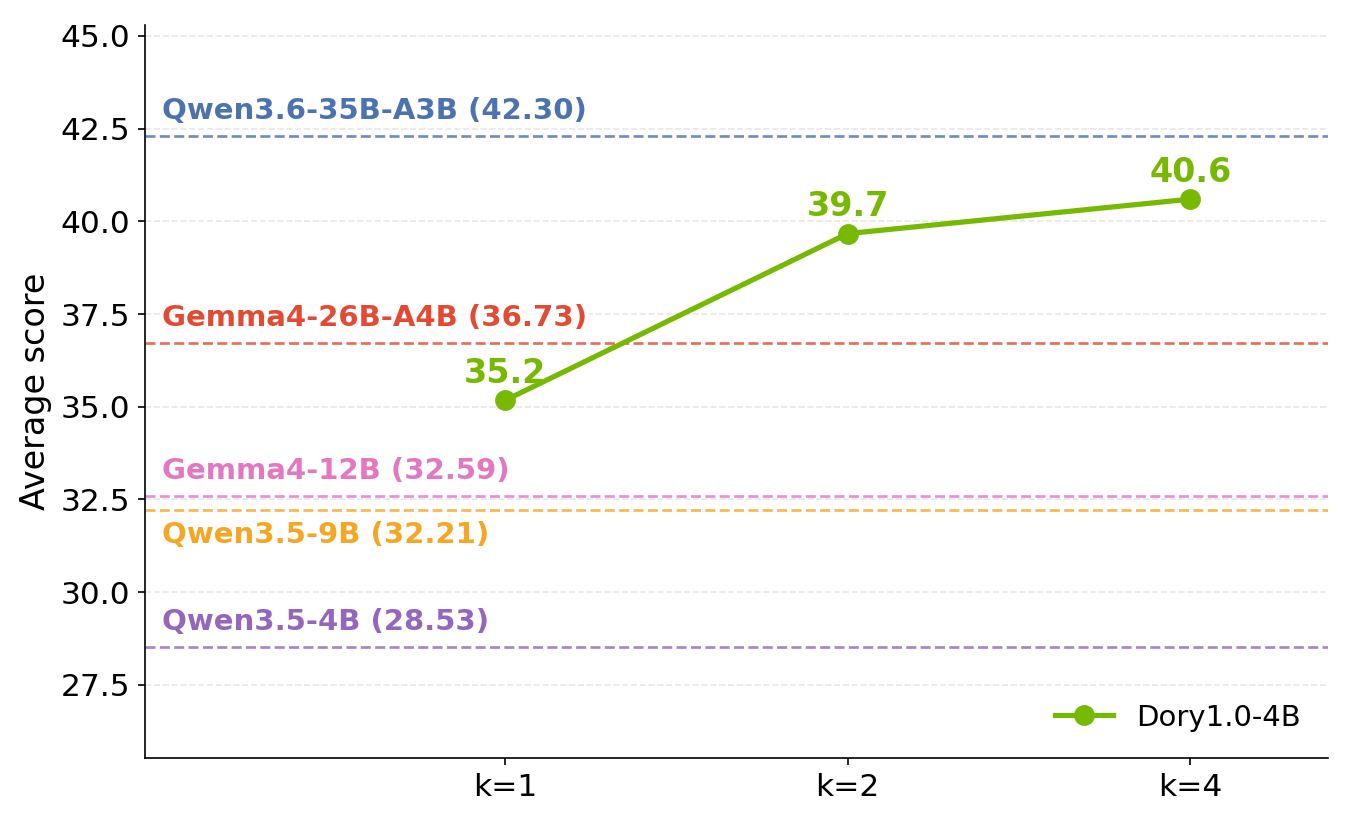}
        \caption{Average score vs.\ loop count $k$.}
        \label{fig:intro_avg}
    \end{subfigure}\hfill
    \begin{subfigure}[t]{0.49\linewidth}
        \centering
        \includegraphics[width=\linewidth]{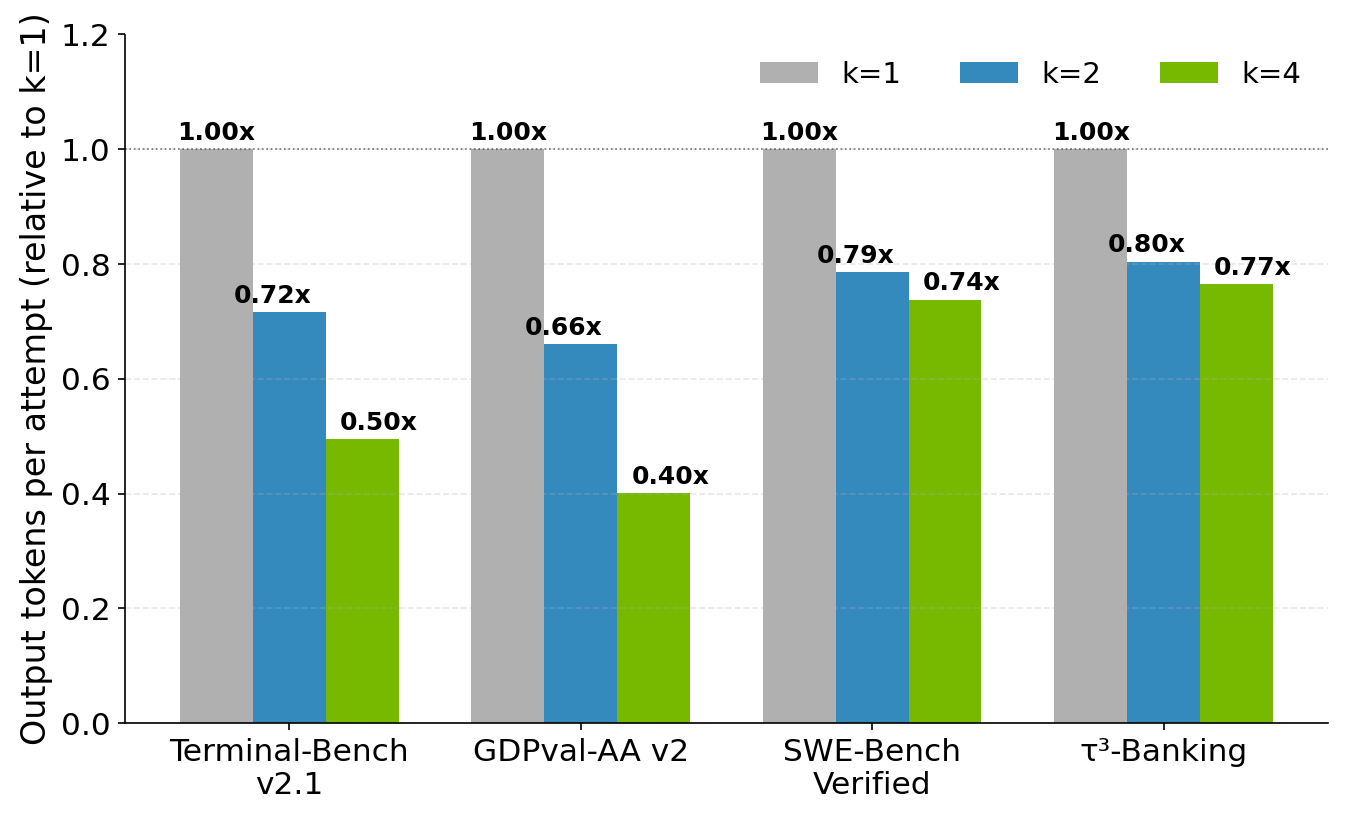}
        \caption{Output tokens per attempt, relative to $k=1$.}
        \label{fig:intro_tokens}
    \end{subfigure}
    \caption{Accuracy and token efficiency of \ourmodel across inference loop counts $k$. (a)~An equal-weight average of 12 benchmarks:
    SciCode, SWE-Bench Verified, Terminal-Bench v2.1, $\tau^3$-Banking, GDPval-AA v2, GPQA Diamond, Humanity's Last Exam (HLE), IFBench, AA-LCR, GDP.pdf, and AA-Omniscience accuracy and non-hallucination, with GDPval-AA v2 Elo normalized as $(\mathrm{Elo}-500)/20$. (b)~Reduction in the number of output tokens needed to solve agentic tasks as the number of loops increases.}
    \label{fig:intro_headline}
\end{figure}

Figure~\ref{fig:intro_headline} summarizes the effect of looping. On the equal-weight average of 12 benchmarks (Table~\ref{tab:posttraining_evals}), \ourmodel improves from 35.2 at one loop to 39.7 at two loops and 40.6 at four loops (Figure~\ref{fig:intro_avg}). It outperforms larger models such as Qwen3.5-9B, Gemma~4 12B, and Gemma~4 26B-A4B, and comes within 1.7 points of Qwen3.6-35B-A3B, a model with roughly 9$\times$ as many total parameters. The additional latent computation also makes \ourmodel more token-efficient: across four agentic benchmarks, four loops reduce the number of generated tokens per attempt by 1.3--2.5$\times$ relative to one loop, with the largest reductions on GDPval-AA v2 and Terminal-Bench v2.1 (Figure~\ref{fig:intro_tokens} and Section~\ref{sec:posttraining_token_efficiency}), partially offsetting the higher compute per token.

\ourmodel uses a normalized Transformer (nGPT) backbone~\citep{loshchilov2025ngptnormalizedtransformerrepresentation} with gated grouped-query attention and a 3:1 interleaving of sliding-window and full-attention layers. Its 72 layers are partitioned into a 16-layer input block, a 40-layer shared recurrent block, and a 16-layer output block. The input and output blocks run once, while the recurrent block runs $k$ times, with the input block's output re-injected before every loop. 

\ourmodel was pretrained on 6T tokens. The model was first pretrained with $k=1$ on $\sim$4.5T tokens as a standard, non-looped model in NVFP4~\citep{nvidia2025pretraininglargelanguagemodels}, which yields an end-to-end speedup of 20\% over BF16. We then introduced looping during a BF16 continuation over $\sim$1.5T tokens, sampling the loop count from $\{1,2,3,4\}$ and training with truncated backpropagation through loops. Finally, we extended the context length to 256K tokens over an additional 300B tokens.

We post-trained \ourmodel with an extended SFT stage followed by a short RLVR stage. SFT ran for nine epochs over packed 256K-token sequences containing both thinking and non-thinking responses, and retained the random loop sampling used in pretraining. RLVR then ran 200 steps of asynchronous GRPO~\citep{grpo} across 23 environments spanning agentic tasks, coding, reasoning, long-context question answering, instruction following, and safety. A key ingredient of RLVR is a difficulty-aware loop curriculum: rollout workers run at fixed loop counts, and prompts with lower profiled pass rates are preferentially routed to deeper loops. In an ablation, this curriculum yields its largest gains at four loops, for example raising Terminal-Bench v2.1 accuracy from 14.0\% to 21.4\%. At four loops, \ourmodel outperforms Gemma~4 26B-A4B on SWE-Bench Verified (61.8\% vs.\ 57.4\%) and GDPval-AA v2 (967 vs.\ 752 Elo; Table~\ref{tab:posttraining_evals}).

With standard per-loop caching, the recurrent block's KV cache grows linearly with $k$. \ourmodel supports last-loop KV caching~\citep{ladhak2026elr}, in which each loop overwrites the previous loop's cache entries, so attention KV memory stays at its one-loop size for any $k$. At four loops, last-loop caching changes benchmark scores by between $-2.1$ and $+0.4$ points relative to per-loop caching, with nearly unchanged output lengths. 

Finally, \ourmodel was converted to MXFP8 using post-training quantization, and to NVFP4 and Q5\_K\_M using quantization-aware distillation~\citep{qadnvfp4inference}. 
All checkpoints are open-sourced under the OpenMDW-1.1 license at \url{https://huggingface.co/collections/nvidia/nemotron-labs-dory}.

\begin{figure*}[h!]
    \centering
    \input{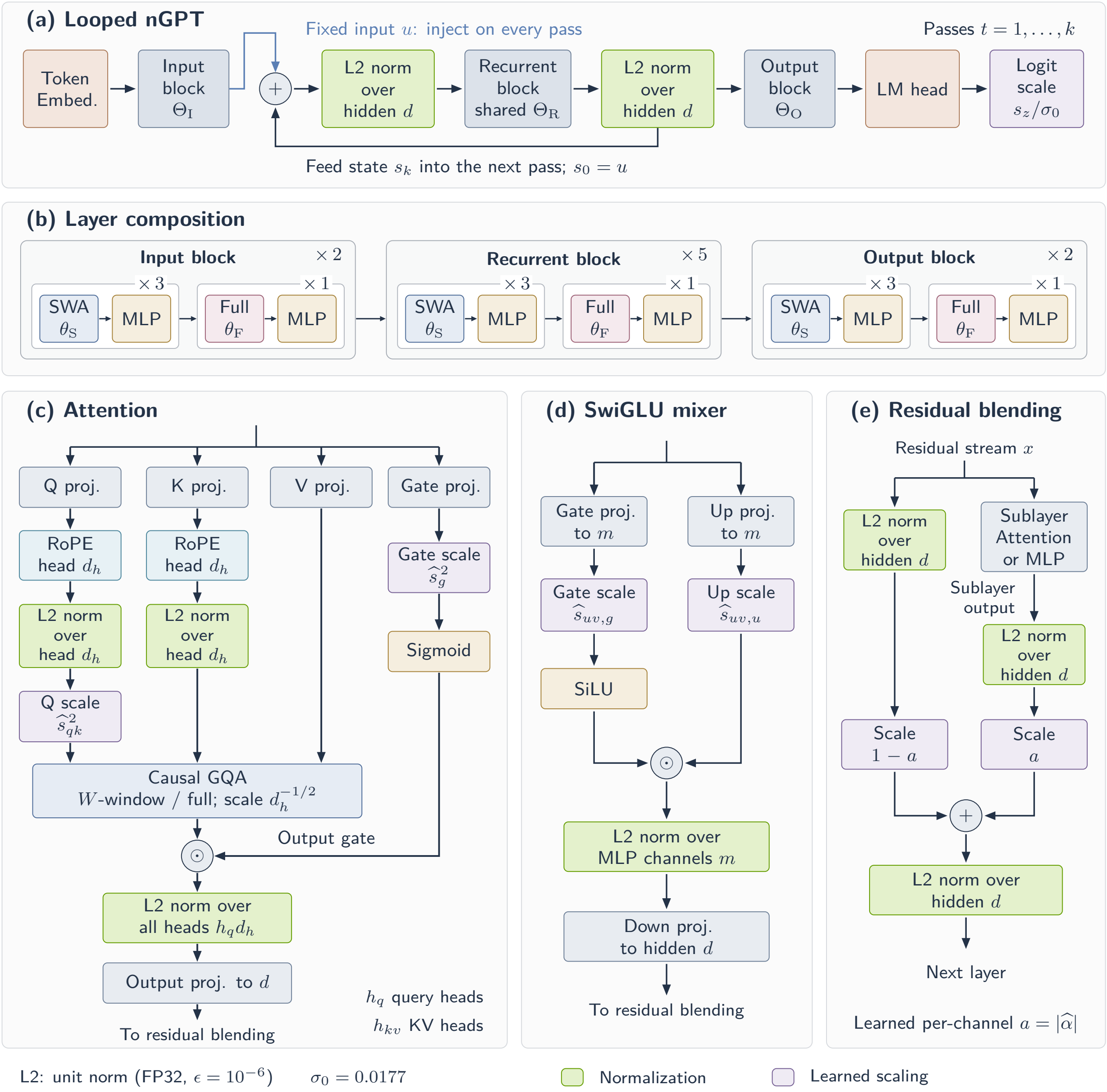}
    \caption{
    \ourmodel's looped nGPT architecture. A 16-layer input block produces a fixed representation $u$.
    A 40-layer recurrent block refines the state over $k$ passes (up to four), sharing all of its weights and learned scales across each loop.
    Each loop adds $u$ to the current state, initialized as $s_0=u$, and applies L2 normalization before and after the recurrent block.
    A 16-layer output block processes the final state before the LM head.
    Within each block, an eight-layer pattern contains three SWA--SwiGLU pairs followed by one full-attention--SwiGLU pair, giving a 3:1 ratio of sliding-window attention (SWA) to full attention. 
    SWA uses a window of $W$=1024 tokens with RoPE base $\theta_{\rm S}=10^4$; full attention uses $\theta_{\rm F}=10^7$. RoPE acts on all Q/K dimensions.
    Each sublayer updates the residual stream by blending its normalized output with the normalized residual using learned per-channel weights, then normalizing the result.
    }
 \label{fig:dory_arch}
  \end{figure*}

\section{Architecture}
\ourmodel is based on the normalized Transformer~\citep{loshchilov2025ngptnormalizedtransformerrepresentation}, retaining its residual-update rule. Our full nGPT implementation additionally L2-normalizes the concatenated, gated attention outputs before the output projection and the SwiGLU activations before the MLP down-projection. We implement these normalizations with fused Triton kernels. We initialize the raw residual-rate, query–key scale, and logit-scale parameters at 0.0177, rather than the paper's $1/\sqrt{d_{\mathrm{model}}}$, while retaining an initial effective residual rate of 0.05.

We augment this formulation with input-dependent sigmoid attention-output gates and grouped-query attention with 16 query heads and 4 KV groups. The stack contains 72 physical layers: 36 MLP-only layers interleaved with 27 sliding-window and 9 full-attention layers. Thus, the attention layers follow a 3:1 sliding-window/full-attention pattern. Figure~\ref{fig:dory_arch} shows the overall architecture of \ourmodel.

\section{Pretraining}
\label{sec:pretraining}
In this section, we provide details on the NVFP4 pretraining recipe, data, and hyperparameters used for \ourbasemodel. We also share details on the long-context extension phase.

\subsection{NVFP4 Training}
We first pretrain a model using Megatron-LM and NVIDIA Transformer Engine's NVFP4 block-scaling recipe \citep{nvidia2025pretraininglargelanguagemodels}. This mixed-precision recipe yields an end-to-end training speedup of approximately 20\% over BF16 by using E2M1 4-bit operands for supported linear-layer matrix multiplications, while retaining BF16 model weights and FP32 optimizer master weights, Adam moments, and gradient accumulation and reduction. We do not exempt the first or last decoder layers from NVFP4. We use distributed Adam with $(\beta_1, \beta_2) = (0.9, 0.95)$, global gradient-norm clipping at 1.0, and no weight decay or learning-rate warmup.

We use the pretraining data of Nemotron~3 Ultra~\citep{nvidia2026nemotron3ultraopen}, following its two-phase curriculum~\citep{feng2024maximizedataspotentialenhancing}: a phase-1 mixture that favors diversity, followed by a phase-2 mixture that favors quality. We train on approximately 3T tokens of phase-1 data at a sequence length of 8,192, then switch to the phase-2 mixture and increase the sequence length to 32,768. Reducing the global batch size from 768 to 192 sequences preserves 6,291,456 tokens per optimizer step. The learning-rate schedule spans both NVFP4 pretraining and the subsequent BF16 continuation: we hold the learning rate at $3.2\times10^{-4}$ through 4.5T tokens, then apply minus-square-root decay toward $10^{-5}$ at 6T tokens. The NVFP4 stage ends at iteration 720,750, after approximately 4.535T tokens.

\subsection{Training with Loops}

We continue from the NVFP4 checkpoint in BF16 for approximately 1.465T additional tokens, reaching the 6T endpoint. This is a normal checkpoint restart: optimizer, scheduler, RNG, and data-position state are preserved, and the phase-2 data mixture remains unchanged. We retain the 32,768-token sequence length and global batch size of 192, using 192 data-parallel GPUs across 48 nodes, a microbatch size of one, and no tensor, pipeline, or context parallelism.

We partition the pretrained stack into a 16-layer input block, a shared 40-layer recurrent block, and a 16-layer output block. The input and output blocks run once, while the recurrent block runs $k$ times with all of its parameters, including nGPT residual rates, shared across invocations. The effective forward depth is $32+40k$, ranging from 72 to 192 physical-layer applications. Let $x$ denote the input block's output, $F_\theta$ the shared recurrent block, and $N$ per-token L2 normalization. We initialize the recurrent state by copying $x$. Our loop-boundary transforms add this same $x$ to the current state before every invocation and replace the state with the recurrent block's output. With normalization at both boundaries, the forward recurrence is

$$
s_0=x,\qquad u_t=N(s_{t-1}+x),\qquad s_t=N(F_\theta(u_t)),\qquad t=1,\ldots,k.
$$

These transforms introduce no learned adapter parameters or outer residual addition; the final state $s_k$ is passed to the output block and language-model head.

We use a base loop-count distribution with probability 0.25 for each $k\in\{1,2,3,4\}$. The draw is deterministic in the optimizer step and microbatch index and is shared across data-parallel replicas, rather than sampled independently on each replica. We additionally set a rank-level override fraction of 0.1: a rotating subset of replicas is forced to execute exactly one loop, while the remaining replicas use the common sampled depth \citep{ladhak2026elr}. Integer rounding selects 19 of the 192 replicas (approximately 9.9\%) for this override. The subset rotates with the step and microbatch index, providing additional one-loop training signal; consequently, the realized depth mixture is not uniform.

\subsection{Truncated Backpropagation}
We use truncated backpropagation through loops (TBPTL) with depth one. Earlier invocations run without gradient tracking; the final recurrent invocation and the input/output blocks remain trainable, with the input block receiving gradients through the final invocation's residual input connection. We apply next-token cross-entropy only after the final sampled invocation, without intermediate-exit or early-loop auxiliary losses.

\FloatBarrier

\subsection{Long-Context Extension}

Starting from the 6T recurrent checkpoint, we perform a long-context extension in BF16, increasing the sequence length from 32,768 to 262,144 tokens. An additional approximately 300B tokens brings the total training budget to approximately 6.3T tokens. We increase the full-attention RoPE base from 500,000 to 10,000,000, while retaining the sliding-window RoPE base of 10,000 and window size of 1,024. The recurrent architecture, input/output transforms, base loop-count distribution, and TBPTL depth remain unchanged. We continue the pretraining objective on a long-context data blend similar to Nemotron 3.5 Super~\citep{nvidia2026nemotron35super}.

Unlike the NVFP4-to-BF16 continuation, we load model weights only and reset the optimizer, scheduler, RNG, and iteration state. We train for 47,683 optimizer steps at a constant learning rate of $10^{-5}$ without warmup. Reducing the global batch size from 192 to 24 sequences preserves 6,291,456 tokens per step. We use 192 GPUs across 48 nodes with eight-way context parallelism, 24-way data parallelism, a microbatch size of one, and no tensor or pipeline parallelism. All-gather communication accommodates the sliding-window attention implementation. With 24 data-parallel replicas, the unchanged 0.1 override setting selects two replicas (approximately 8.3\%) for one-loop execution after integer rounding.

\subsection{Base Model Evaluations}
\label{subsec:base_model_evals}

We evaluate \ourbasemodel after the long-context extension. The base model has seen approximately 6.3T training
tokens. Table~\ref{tab:base-model-comparison} reports ten benchmarks at fixed inference loop counts
$k\in\{1,2,4,6,8\}$. Here, $k$ is the number of passes of the shared
40-layer recurrent block. All columns use the same model weights, with training-time 
rank-based loop overrides disabled. Training samples one to four loops; the six- and
eight-loop evaluations measure inference-depth extrapolation.

\begin{table}[t!]
\centering
\small
\setlength{\tabcolsep}{6pt}
\renewcommand{\arraystretch}{1.15}
\begin{adjustbox}{max width=0.95\textwidth}
\begin{tabular}{lr|ccccc}
\toprule
\textbf{Task} & \textbf{Metric} &
\multicolumn{5}{c}{\textbf{\ourbasemodel}} \\
\cmidrule(lr){3-7}
& \textbf{Inference loops} & \textbf{1} & \textbf{2} & \textbf{4} & \textbf{6} & \textbf{8} \\
\rowcolor{black!5}
\multicolumn{7}{l}{\textbf{General Knowledge}} \\
MMLU (Str) & \textit{5-shot EM} & 68.20 & 70.77 & 71.19 & 71.22 & 71.17 \\
MMLU-Pro & \textit{5-shot CoT EM} & 51.52 & 55.06 & 56.09 & 55.72 & 55.86 \\
\midrule

\rowcolor{black!5}
\multicolumn{7}{l}{\textbf{Mathematical Reasoning}} \\
GSM8K & \textit{8-shot CoT EM} & 86.43 & 89.99 & 89.84 & 89.01 & 89.99 \\
MATH-500 & \textit{4-shot greedy EM} & 60.60 & 65.00 & 69.40 & 68.40 & 69.40 \\
\midrule

\rowcolor{black!5}
\multicolumn{7}{l}{\textbf{Code Generation}} \\
HumanEval & \textit{0-shot pass@1, $n=64$} & 58.33 & 59.76 & 62.34 & 60.52 & 61.90 \\
MBPP-Sanitized & \textit{3-shot pass@1, $n=30$} & 57.09 & 61.50 & 60.89 & 59.13 & 61.73 \\
\midrule

\rowcolor{black!5}
\multicolumn{7}{l}{\textbf{Commonsense Reasoning}} \\
ARC-Challenge (Llama) & \textit{25-shot acc\_norm} & 87.63 & 89.08 & 88.74 & 89.85 & 88.91 \\
HellaSwag & \textit{10-shot acc\_norm} & 78.52 & 79.53 & 79.76 & 79.75 & 79.91 \\
WinoGrande & \textit{5-shot acc} & 74.11 & 76.24 & 77.43 & 77.35 & 77.82 \\
\midrule

\rowcolor{black!5}
\multicolumn{7}{l}{\textbf{Reading Comprehension}} \\
RACE-H$^{\dagger}$ & \textit{0-shot acc} & 83.06 & 85.26 & 86.03 & 86.03 & 85.74 \\
\bottomrule
\end{tabular}
\end{adjustbox}
\caption[Base-model evaluation results by inference loop count]{Base-model evaluation results for \textbf{\ourbasemodel} at fixed inference loop counts. All scores are percentages, rounded to two decimal places; higher is better. EM denotes the task's exact-match metric, and acc\_norm denotes length-normalized multiple-choice accuracy. For code tasks, $n$ is the number of sampled completions per problem used to estimate pass@1. MATH-500 uses one greedy completion, not sampled pass@1. $^{\dagger}$RACE-H uses the last-question-per-passage protocol described below. As performance tends to saturate after four loops, we restrict post-training results to at most four loops.}
\label{tab:base-model-comparison}
\end{table}

\paragraph{Evaluation implementation.}
We use NVIDIA's fork of the LM Evaluation Harness%
\footnote{Upstream harness: \url{https://github.com/EleutherAI/lm-evaluation-harness}.}
through the Megatron-LM completions server. The model receives raw base-model
prompts without a chat template or system instruction.
All evaluations use the full test or validation split.

\begin{enumerate}
\setlength{\itemsep}{0.4em}

\item \textbf{General knowledge.}
MMLU~\citep{hendrycks2021measuringmassivemultitasklanguage} uses the
string-generation protocol: five demonstrations, greedy decoding capped
at two tokens, and exact matching of the extracted answer letter.
We report the question-weighted aggregate across 57 subjects and 14,042
test questions, not the subject-macro average or a likelihood-based MMLU
score. MMLU-Pro~\citep{wang2024mmluprorobustchallengingmultitask} uses five-shot chain-of-thought prompting on 12,032
questions, greedy decoding capped at 1,024 tokens, and exact match on the
extracted answer.

\item \textbf{Mathematical reasoning.}
GSM8K~\citep{cobbe2021trainingverifierssolvemath} uses eight-shot
chain-of-thought prompting over 1,319 test problems. MATH-500~\citep{hendrycks2021measuringmathematicalproblemsolving,lightman2023lets} uses four
fixed demonstrations over 500 test problems. Both use one greedy
completion per problem, a 1,024-token generation limit, and exact match
after answer extraction.

\item \textbf{Code generation.}
HumanEval~\citep{chen2021evaluatinglargelanguagemodels} uses zero-shot
prompts for 164 problems and 64 completions per problem, with a 512-token
generation limit. MBPP-Sanitized~\citep{austin2021programsynthesislargelanguage}
uses three demonstrations for each of 257 test problems and 30 completions
per problem, with a 768-token generation limit. Both sample at temperature
0.6 and top-$p=0.95$. We apply the EvalPlus
sanitizer~\citep{Liu_Is_Your_Code_2023}, but execute the original HumanEval
or sanitized-MBPP tests, not the additional HumanEval+ or MBPP+ tests.
Pass@1 is the mean fraction of passing completions per problem, not the
fraction of problems with at least one passing sample.

\item \textbf{Commonsense reasoning.}
We evaluate ARC-Challenge~\citep{Clark2018ThinkYH} using the Llama-style
25-shot task on 1,172 test questions, HellaSwag%
~\citep{zellers2019hellaswagmachinereallyfinish} using ten-shot prompting
on 10,042 validation examples, and WinoGrande%
~\citep{sakaguchi2019winograndeadversarialwinogradschema} using five-shot
prompting on 1,267 validation examples. These are multiple-choice
likelihood evaluations. We report \texttt{acc\_norm} for ARC-Challenge
and HellaSwag, and \texttt{acc} for WinoGrande.

\item \textbf{Reading comprehension.}
RACE-H~\citep{lai-etal-2017-race} uses the high-school test split with
1,045 passage-level examples. Only the final question of each passage is
scored, with the preceding questions and their gold answers included in
the prompt and no additional demonstrations. Scores are therefore not
directly comparable to all-question RACE results.

\end{enumerate}

\section{Post-Training}
\label{sec:posttraining}

\begin{figure}[!h]
    \centering
    \definecolor{dorypipelinegreen}{RGB}{118,185,0}
    \resizebox{0.98\linewidth}{!}{%
    \begin{tikzpicture}[
        stage/.style={
            draw=dorypipelinegreen!85!black,
            fill=dorypipelinegreen!14,
            rounded corners=7pt,
            line width=0.9pt,
            minimum height=1.25cm,
            align=center,
            inner sep=8pt,
            font=\small
        },
        annotation/.style={font=\small, text=black!70, anchor=north, inner sep=0pt},
        flow/.style={->, >=stealth, line width=1.2pt, draw=dorypipelinegreen!70!black}
    ]
        \node[stage, minimum width=2.4cm] (base) {
            \textbf{Base}\\[4pt]
            \mbox{256K context}
        };
        \node[stage, minimum width=5.4cm, right=0.65cm of base] (sft) {
            \textbf{SFT}\\[4pt]
            \mbox{4.15M packed sequences}
        };
        \node[stage, minimum width=5.0cm, right=0.65cm of sft] (rlvr) {
            \textbf{RLVR}\\[4pt]
            \mbox{Light GRPO $\cdot$ 200 steps}
        };
        \node[stage, fill=dorypipelinegreen!28, minimum width=2.4cm,
              right=0.65cm of rlvr] (model) {
            \mbox{\textbf{\ourmodel{} (BF16)}}
        };
        \node[annotation] at ([yshift=-8pt]sft.south) {Random Loop Sampling};
        \node[annotation] at ([yshift=-8pt]rlvr.south) {Difficulty-Aware Loop Curriculum};
        \draw[flow] (base.east) -- (sft.west);
        \draw[flow] (sft.east) -- (rlvr.west);
        \draw[flow] (rlvr.east) -- (model.west);
    \end{tikzpicture}%
    }
    \caption{Overview of the post-training pipeline for \ourmodel.
    The base context window is 262,144 tokens (256K). SFT trains on packed sequences of 262,144 tokens, and samples loop counts for each microbatch randomly, while RLVR
    routes harder prompts to deeper loops with higher probability.}
    \label{fig:posttraining_pipeline}
\end{figure}
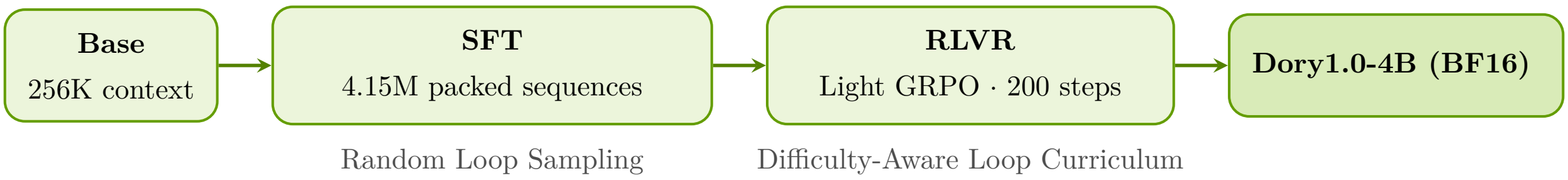

Our post-training pipeline (Figure~\ref{fig:posttraining_pipeline}) combines a heavy supervised fine-tuning (SFT) stage and a light 200-step reinforcement learning with verifiable rewards (RLVR) stage. Initialized from the long-context extended checkpoint, SFT adapts \ourmodel~to instruction following using mixed thinking and non-thinking responses. The RLVR stage begins from the 9-epoch SFT checkpoint and optimizes the model using task-specific rewards across multiple environments. Both stages train the shared recurrent block at multiple loop counts. While SFT maintains the variable-loop setup from pretraining and context extension, RLVR dynamically routes harder prompts (lower pass rates) to deeper loops with higher probability.

\subsection{Supervised Fine-Tuning}
\label{sec:sft}
Starting from the long-context extended checkpoint, we perform SFT in BF16 using Megatron-LM on packed sequences of 262,144 tokens with a global batch size of 128 and a microbatch size of one. We retain the optimizer configuration used during long-context extension: distributed Adam with $(\beta_1,\beta_2)=(0.9,0.95)$, a constant learning rate of $10^{-5}$ without warmup, no weight decay, and gradient-norm clipping at 1.0. As in long-context extension, SFT uses eight-way context parallelism, with no tensor or pipeline parallelism. Training runs on 32 nodes with 128 GB300 GPUs and a one-loop override fraction of 0.1 on top of uniform loop-count distribution for $k\in\{1,2,3,4\}$. We use TBPTL with depth one, as in pretraining. We select the SFT checkpoint at iteration 32,400 to initialize RLVR.

The SFT data blend is the same as the text blend for post-training Nemotron 3.5 Super,\footnote{\url{https://huggingface.co/nvidia/NVIDIA-Nemotron-3.5-Super-VL-120B-A12B-BF16}} and packed offline to a maximum sequence length of 262,144 tokens. The blend contains both thinking and non-thinking responses. For thinking examples, the prompt supplies an open \texttt{<think>} tag, and the model generates a reasoning trace followed by \texttt{</think>} and the final response. For non-thinking examples, the prompt supplies an empty \texttt{<think></think>} block, and the model generates the response directly. The chat template delimits messages with \texttt{<|im\_start|>} and \texttt{<|im\_end|>} and exposes tool definitions in the system message. Tool calls use XML-style \texttt{<tool\_call>} blocks while tool results are wrapped in \texttt{<tool\_response>} blocks within user-formatted messages. We optimize token-level cross-entropy on assistant tokens, including the reasoning trace and closing \texttt{</think>} tag in thinking examples, while masking system, user, and tool tokens from the loss.

Extended SFT yields continued improvements across multiple benchmarks evaluated with one loop. Between iterations 3,600 and 32,400 (epochs one and nine), SciCode improves from 13.11\% to 18.14\%, Terminal-Bench v2.1 from 4.2\% to 13.5\%, and GPQA Diamond from 43.13\% to 54.95\%. Over the same interval, IFBench increases from 28.94\% to 38.50\%, MATH-500 from 60.2\% to 75.2\%, and AA-LCR from 29.5\% to 37.3\%. These results show that longer SFT continues to improve coding, agentic, reasoning, and instruction-following performance at a fixed one-loop inference budget.

\subsection{RLVR} 

\paragraph{Environments and rewards.} We train across 23 single-step environments covering coding, agentic tasks, reasoning, long-context question answering, instruction following, structured outputs, safety, and other environments. For LLM-based reward, we use Qwen3-235B-A22B-Instruct-2507\footnote{\url{https://huggingface.co/Qwen/Qwen3-235B-A22B-Instruct-2507}} for scoring the policy responses.

\paragraph{Reward profiling and data selection.} Before RLVR training, we profile candidate prompts using the SFT checkpoint at iteration 32,400 with one-loop execution. For each prompt, eight responses are collected and graded with the corresponding environment's reward function. We retain prompts whose mean reward is strictly between 0 and 1, excluding extremes of consistently zero reward and consistently full reward. Each prompt's profiled pass rate (or mean profiling reward) is preserved for the \emph{difficulty-aware loop curriculum} introduced later. We then re-balance the profiled data by oversampling targeted environments and filter prompts to fit the RLVR training context. The final RLVR data mixture comprises $\sim$47\% agentic tasks, $\sim$27.5\% reasoning and coding, $\sim$8.1\% long-context question answering, $\sim$17.4\% instruction following, safety, and other domains. 

\paragraph{RLVR training setup.} Starting from the SFT checkpoint, we train for 200 steps using asynchronous group relative policy optimization (GRPO)~\citep{grpo} in NeMo RL~\citep{nemo-rl}. Each step samples 384 prompts and generates 16 responses per prompt, yielding a global batch size of 6,144 rollouts. We train in BF16 with distributed Adam, using $(\beta_1,\beta_2)=(0.9,0.999)$, a constant learning rate of $4\times10^{-6}$ without warmup, no weight decay, and gradient-norm clipping at 1.0. Rollout workers use temperature 1.0 and top-$p=1.0$ within the 65,536-token context window. We launch separate vLLM rollout workers each with a fixed loop count $k$ from $\{1,2,3,4\}$. Within each global batch, prompt quotas follow weights of 4:1:1:1 for $k=1,2,3,4$, respectively. Each prompt is assigned a loop count before rollout expansion, and all 16 rollouts in its GRPO group use that same count.

\begin{figure}[t]
    \centering
    \definecolor{dorycurriculumgreen}{RGB}{118,185,0}
    \resizebox{0.98\linewidth}{!}{%
    \begin{tikzpicture}[
        font=\small,
        bar/.style={draw=dorycurriculumgreen!85!black, line width=0.5pt},
        axis/.style={->, >=stealth, draw=black!65, line width=0.7pt},
        flow/.style={->, >=stealth, draw=dorycurriculumgreen!70!black,
                     line width=1.2pt}
    ]
        \foreach \x/\title/\ylabel in {
            0/{Profiled prompt difficulty}/{Prompt count},
            4.9/{Uniform sampling}/{Density},
            10.5/{Difficulty-aware loop routing}/{Prompt count}
        } {
            \begin{scope}[xshift=\x cm]
                \node at (1.55,2.7) {\textbf{\title}};
                \draw[axis] (-0.08,0) -- (3.3,0);
                \draw[axis] (0,-0.03) -- (0,2.05);
                \node[rotate=90, font=\scriptsize, text=black!70]
                    at (-0.33,1.05) {\ylabel};
            \end{scope}
        }
        \node[font=\scriptsize, text=black!70] at (1.55,2.2)
            {Pass rate $\longrightarrow q_i$; weight $\alpha$};
        \node[font=\scriptsize, text=black!70] at (6.45,2.2)
            {$u_i\sim\mathrm{Uniform}(0,1)$; weight $1-\alpha$};
        \node[font=\scriptsize, text=black!70] at (12.05,2.2)
            {\mbox{Color indicates loop count}};

        \draw[bar, fill=dorycurriculumgreen!28] (0.0180,0) rectangle (0.4249,1.8000);
        \draw[bar, fill=dorycurriculumgreen!14] (10.5180,0.0000) rectangle (10.9249,0.1601);
        \draw[bar, fill=dorycurriculumgreen!32] (10.5180,0.1601) rectangle (10.9249,0.6096);
        \draw[bar, fill=dorycurriculumgreen!52] (10.5180,0.6096) rectangle (10.9249,1.0610);
        \draw[bar, fill=dorycurriculumgreen!78] (10.5180,1.0610) rectangle (10.9249,1.8000);
        \draw[bar, fill=dorycurriculumgreen!28] (0.4609,0) rectangle (0.8677,1.4437);
        \draw[bar, fill=dorycurriculumgreen!14] (10.9609,0.0000) rectangle (11.3677,0.2674);
        \draw[bar, fill=dorycurriculumgreen!32] (10.9609,0.2674) rectangle (11.3677,0.6283);
        \draw[bar, fill=dorycurriculumgreen!52] (10.9609,0.6283) rectangle (11.3677,0.9896);
        \draw[bar, fill=dorycurriculumgreen!78] (10.9609,0.9896) rectangle (11.3677,1.4437);
        \draw[bar, fill=dorycurriculumgreen!28] (0.9037,0) rectangle (1.3106,1.1438);
        \draw[bar, fill=dorycurriculumgreen!14] (11.4037,0.0000) rectangle (11.8106,0.3016);
        \draw[bar, fill=dorycurriculumgreen!32] (11.4037,0.3016) rectangle (11.8106,0.5861);
        \draw[bar, fill=dorycurriculumgreen!52] (11.4037,0.5861) rectangle (11.8106,0.8721);
        \draw[bar, fill=dorycurriculumgreen!78] (11.4037,0.8721) rectangle (11.8106,1.1438);
        \draw[bar, fill=dorycurriculumgreen!28] (1.3466,0) rectangle (1.7534,0.9375);
        \draw[bar, fill=dorycurriculumgreen!14] (11.8466,0.0000) rectangle (12.2534,0.3043);
        \draw[bar, fill=dorycurriculumgreen!32] (11.8466,0.3043) rectangle (12.2534,0.5392);
        \draw[bar, fill=dorycurriculumgreen!52] (11.8466,0.5392) rectangle (12.2534,0.7720);
        \draw[bar, fill=dorycurriculumgreen!78] (11.8466,0.7720) rectangle (12.2534,0.9375);
        \draw[bar, fill=dorycurriculumgreen!28] (1.7894,0) rectangle (2.1963,0.7875);
        \draw[bar, fill=dorycurriculumgreen!14] (12.2894,0.0000) rectangle (12.6963,0.2959);
        \draw[bar, fill=dorycurriculumgreen!32] (12.2894,0.2959) rectangle (12.6963,0.4939);
        \draw[bar, fill=dorycurriculumgreen!52] (12.2894,0.4939) rectangle (12.6963,0.6916);
        \draw[bar, fill=dorycurriculumgreen!78] (12.2894,0.6916) rectangle (12.6963,0.7875);
        \draw[bar, fill=dorycurriculumgreen!28] (2.2323,0) rectangle (2.6391,0.6562);
        \draw[bar, fill=dorycurriculumgreen!14] (12.7323,0.0000) rectangle (13.1391,0.2760);
        \draw[bar, fill=dorycurriculumgreen!32] (12.7323,0.2760) rectangle (13.1391,0.4392);
        \draw[bar, fill=dorycurriculumgreen!52] (12.7323,0.4392) rectangle (13.1391,0.6026);
        \draw[bar, fill=dorycurriculumgreen!78] (12.7323,0.6026) rectangle (13.1391,0.6562);
        \draw[bar, fill=dorycurriculumgreen!28] (2.6751,0) rectangle (3.0820,0.4312);
        \draw[bar, fill=dorycurriculumgreen!14] (13.1751,0.0000) rectangle (13.5820,0.1947);
        \draw[bar, fill=dorycurriculumgreen!32] (13.1751,0.1947) rectangle (13.5820,0.3037);
        \draw[bar, fill=dorycurriculumgreen!52] (13.1751,0.3037) rectangle (13.5820,0.4112);
        \draw[bar, fill=dorycurriculumgreen!78] (13.1751,0.4112) rectangle (13.5820,0.4312);

        \foreach \j in {0,...,9} {
            \draw[bar, fill=dorycurriculumgreen!20]
                ({4.9+0.31*\j+0.015},0)
                rectangle ({4.9+0.31*(\j+1)-0.015},1.15);
        }
        \foreach \x in {0,10.5} {
            \node[anchor=north, font=\scriptsize, text=black!70]
                at ({\x+0.22},-0.08) {$1/8$};
            \node[anchor=north, font=\scriptsize, text=black!70]
                at ({\x+2.88},-0.08) {$7/8$};
            \node at ({\x+1.55},-0.65) {loop-1 pass rate};
            \node[font=\scriptsize, text=black!70] at ({\x+1.55},-1.02)
                {Harder $\longrightarrow$ Easier};
        }
        \node[anchor=north, font=\scriptsize, text=black!70]
            at (4.9,-0.08) {$0$};
        \node[anchor=north, font=\scriptsize, text=black!70]
            at (8,-0.08) {$1$};
        \node at (6.45,-0.65) {Random score $u_i$};
        \node[font=\Large, text=dorycurriculumgreen!70!black]
            at (4.0,1.0) {$+$};
        \draw[flow] (8.6,1.0) -- (10.0,1.0);
        \node[font=\scriptsize, text=black!70]
            at (9.2,1.6) {Rank $s_i$};
        \node[font=\scriptsize, text=black!70]
            at (9.2,0.4) {Fill quotas};

        \foreach \k/\shade [count=\j from 0] in {1/14,2/32,3/52,4/78} {
            \draw[bar, fill=dorycurriculumgreen!\shade]
                ({3.3+1.8*\j},-1.7) rectangle ({3.65+1.8*\j},-1.45);
            \node[anchor=west] at ({3.75+1.8*\j},-1.575) {$k=\k$};
        }
        \node[text=black!70] at (6.9,-2.2)
            {Lower pass rate $\longrightarrow$ higher probability of deeper loops};
    \end{tikzpicture}%
    }
    \caption{Illustration of difficulty-aware loop routing at $\alpha=0.3$.
    A generic distribution of loop-1 pass rates supplies the
    within-batch midpoint quantiles $q_i$, which are combined with uniform
    sampling $u_i$. After sorting the resulting scores, lower scores fill deeper
    loop slots first. The output histogram preserves the input prompt counts
    and shows the assigned loop counts within each difficulty bin: darker
    segments denote more loop counts. }
    \label{fig:loop_curriculum_interpolation}
\end{figure}
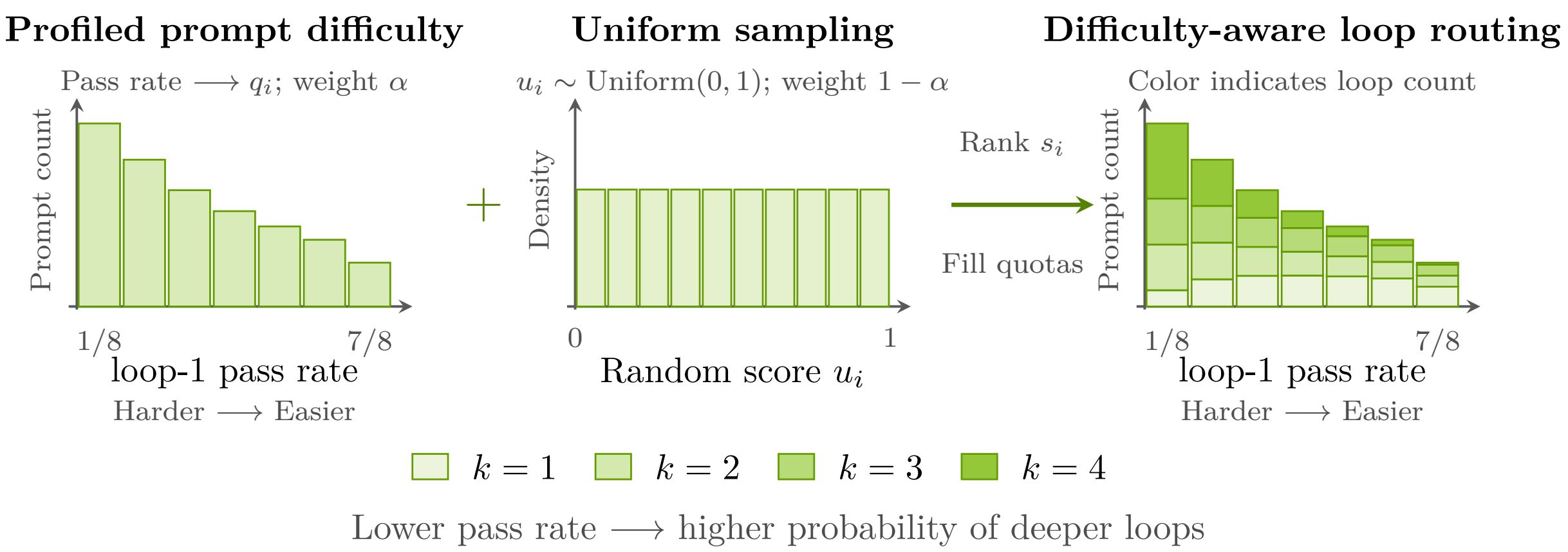

\paragraph{RLVR with difficulty-aware loop curriculum.} The mean profiling reward of each prompt from the reward-profiling stage serves as a proxy for ``difficulty.''
In the RLVR stage, we adopt a difficulty-aware loop-routing curriculum, where ``easier'' prompts are routed with higher probability to rollout workers with fewer loops. Intuitively, two routing strategies are interpolated: (a) ordering prompts by difficulty so that harder prompts receive deeper execution, and (b) assigning prompts independently of their pass rates. Within each prompt batch, we compute the midpoint quantile $q_i$ of each prompt's profiled score, assigning the same quantile to tied values. We interpolate this quantile with a uniform random draw $u_i$ to form a routing score,
\begin{equation}
    s_i = \alpha q_i + (1-\alpha)u_i, \qquad u_i\sim\mathrm{Uniform}(0,1).
\end{equation}
Prompts with lower routing scores fill deeper-loop slots first. At $\alpha=1$, routing follows difficulty order, with random tie-breaking; at $\alpha=0$, it is a uniformly random assignment while obeying the specified prompt loop distribution (e.g., 4:1:1:1 in our case). We use $\alpha=0.3$ to bias routing toward deeper execution for harder prompts while retaining substantial randomness. 
Figure~\ref{fig:loop_curriculum_interpolation} illustrates how the interpolation changes prompt assignments while preserving the loop quotas.

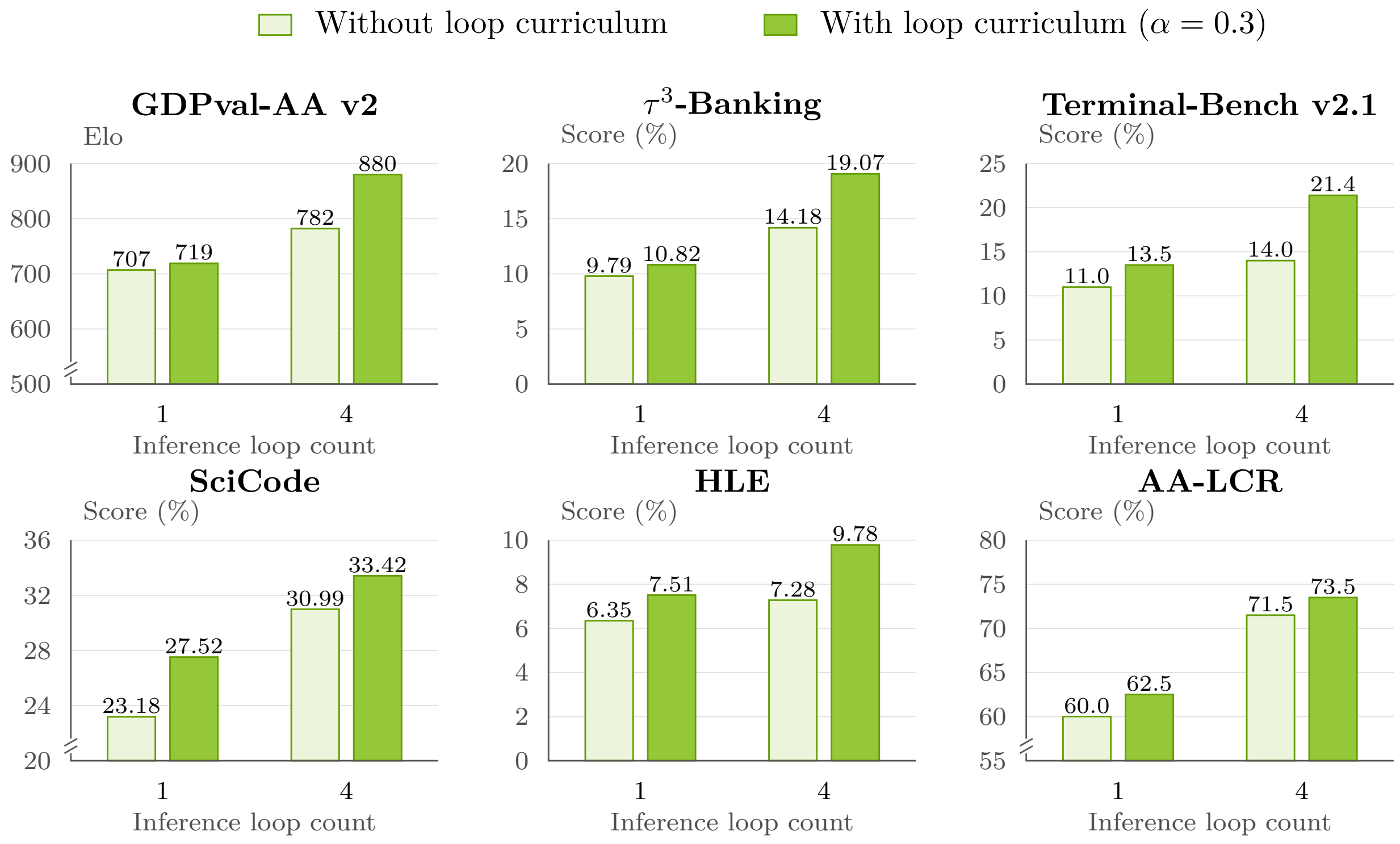
\begin{figure}[htbp]
    \centering
    \definecolor{doryablationgreen}{RGB}{118,185,0}
    \newcommand{\doryablationpanel}[6]{%
        \node[font=\small\bfseries, align=center] at (2,3.05) {#1};
        \node[font=\scriptsize, anchor=west, text=black!70]
            at (0,2.7) {#2};
        \foreach \tick in {#5} {
            \pgfmathsetmacro{\tickheight}{2.4*(\tick-#3)/(#4-#3)}
            \draw[black!12, line width=0.3pt]
                (0,\tickheight) -- (4,\tickheight);
            \node[anchor=east, font=\scriptsize, text=black!70]
                at (-0.08,\tickheight) {\tick};
        }
        \foreach \loops/\without/\with [count=\groupindex] in {#6} {
            \pgfmathsetmacro{\groupx}{1+2*(\groupindex-1)}
            \pgfmathsetmacro{\withoutheight}{2.4*(\without-#3)/(#4-#3)}
            \pgfmathsetmacro{\withheight}{2.4*(\with-#3)/(#4-#3)}
            \draw[ablationbar, fill=doryablationgreen!14]
                (\groupx-0.60,0) rectangle (\groupx-0.08,\withoutheight);
            \draw[ablationbar, fill=doryablationgreen!78]
                (\groupx+0.08,0) rectangle (\groupx+0.60,\withheight);
            \node[anchor=south, font=\fontsize{7}{8}\selectfont, inner sep=1pt]
                at (\groupx-0.34,\withoutheight) {\without};
            \node[anchor=south, font=\fontsize{7}{8}\selectfont, inner sep=1pt]
                at (\groupx+0.34,\withheight) {\with};
            \node[font=\scriptsize, anchor=north]
                at (\groupx,-0.10) {\loops};
        }
        \draw[black!65, line width=0.5pt] (0,2.4) -- (0,0) -- (4,0);
        \ifnum#3>0\relax
            \draw[white, line width=3pt] (0,0.08) -- (0,0.24);
            \draw[black!65, line width=0.5pt]
                (-0.07,0.08) -- (0.07,0.16)
                (-0.07,0.16) -- (0.07,0.24);
        \fi
        \node[font=\scriptsize, anchor=north, text=black!70]
            at (2,-0.43) {Inference loop count};
    }
    \resizebox{0.98\linewidth}{!}{%
    \begin{tikzpicture}[
        ablationbar/.style={draw=doryablationgreen!85!black, line width=0.5pt}
    ]
        \draw[ablationbar, fill=doryablationgreen!14]
            (2.05,3.8) rectangle (2.40,4.02);
        \node[anchor=west, font=\small] at (2.52,3.91)
            {Without loop curriculum};
        \draw[ablationbar, fill=doryablationgreen!78]
            (7.55,3.8) rectangle (7.90,4.02);
        \node[anchor=west, font=\small] at (8.02,3.91)
            {With loop curriculum ($\alpha=0.3$)};

        \begin{scope}
            \doryablationpanel{GDPval-AA v2}{Elo}{500}{900}{500,600,700,800,900}
                {1/707/719,4/782/880}
        \end{scope}
        \begin{scope}[xshift=5.2cm]
            \doryablationpanel{$\tau^3$-Banking}{Score (\%)}{0}{20}{0,5,10,15,20}
                {1/9.79/10.82,4/14.18/19.07}
        \end{scope}
        \begin{scope}[xshift=10.4cm]
            \doryablationpanel{Terminal-Bench v2.1}{Score (\%)}{0}{25}{0,5,10,15,20,25}
                {1/11.0/13.5,4/14.0/21.4}
        \end{scope}
        \begin{scope}[yshift=-4.1cm]
            \doryablationpanel{SciCode}{Score (\%)}{20}{36}{20,24,28,32,36}
                {1/23.18/27.52,4/30.99/33.42}
        \end{scope}
        \begin{scope}[xshift=5.2cm, yshift=-4.1cm]
            \doryablationpanel{HLE}{Score (\%)}{0}{10}{0,2,4,6,8,10}
                {1/6.35/7.51,4/7.28/9.78}
        \end{scope}
        \begin{scope}[xshift=10.4cm, yshift=-4.1cm]
            \doryablationpanel{AA-LCR}{Score (\%)}{55}{80}{55,60,65,70,75,80}
                {1/60.0/62.5,4/71.5/73.5}
        \end{scope}
    \end{tikzpicture}%
    }
    \caption{Loop-curriculum ablation with an early SFT checkpoint.}
    \label{fig:loop_curriculum_ablation}
\end{figure}

\noindent We compare RLVR with and without the difficulty-aware loop curriculum using an earlier SFT checkpoint. Both runs use the same 4:1:1:1 prompt loop distribution; the curriculum run uses $\alpha=0.3$ (Figure~\ref{fig:loop_curriculum_ablation}). The loop curriculum improves all six benchmarks at both one and four loops, with the largest gains on agentic tasks at four loops: GDPval-AA v2 improves by 98 Elo, $\tau^3$-Banking by 4.9 points, and Terminal-Bench v2.1 by 7.4 points.

\subsection{Evaluation}

We evaluate \ourmodel~on benchmarks spanning agentic capabilities, reasoning and knowledge, instruction following, factual reliability, and long-context understanding. Our evaluation follows a recipe similar to that of Nemotron 3.5 Super and Lightning~\citep{nvidia2026nemotron35super}. Evaluation runs are orchestrated through NeMo Evaluator.\footnote{\url{https://github.com/NVIDIA-NeMo/Evaluator}} Most benchmarks use NeMo Gym;\footnote{\url{https://github.com/NVIDIA-NeMo/Gym}} Terminal-Bench v2.1 and SWE-Bench Verified use the evaluator backend with Harbor\footnote{\url{https://github.com/harbor-framework/harbor}} sandboxes on AWS ECS. For benchmarks with small task sets, we evaluate multiple independent responses per task to reduce sampling variability and obtain more reliable score estimates.

Unless otherwise specified, all models in Table~\ref{tab:posttraining_evals} are evaluated using an identical evaluation protocol for each benchmark, including the evaluation harness, task configuration, and scoring procedure. These benchmark scores may differ from those published by Artificial Analysis. The shared protocol makes the benchmark scores and averages comparable across models within our evaluation; externally reported results are explicitly marked.

\begin{table}[!h]
\centering
\small
\setlength{\tabcolsep}{6pt}
\renewcommand{\arraystretch}{1.3}
\resizebox{0.99\textwidth}{!}{%
\begin{tabular}{l|ccc|ccccc}
\toprule
\textbf{Task} & \multicolumn{3}{c|}{\textbf{\ourmodel}} &
\multirow{2}{*}{\makecell{\textbf{Qwen3.5}\\\textbf{4B}}} &
\multirow{2}{*}{\makecell{\textbf{Qwen3.5}\\\textbf{9B}}} &
\multirow{2}{*}{\makecell{\textbf{Gemma 4}\\\textbf{12B}}} &
\multirow{2}{*}{\makecell{\textbf{Gemma 4}\\\textbf{26B-A4B}}} &
\multirow{2}{*}{\makecell{\textbf{Qwen3.6}\\\textbf{35B-A3B}}} \\
\cmidrule(lr){2-4}
& \textbf{1 loop} & \textbf{2 loops} & \textbf{4 loops} & & & & & \\
\midrule
\rowcolor{black!5}
\multicolumn{9}{l}{\textbf{Coding \& Agentic}} \\
SciCode & 27.3 & 34.94 & 35.11 & 25.82 & 28.21 & 37.89 & 40.32 & 37.24 \\
SWE-Bench Verified & 50.3 & 59.5 & 61.8 & 54.7 & 61.0 & 47.3 & 57.40 & 73.4$^{\dagger}$ \\
Terminal-Bench v2.1 & 12.5 & 18.1 & 22.2 & 15 & 24 & 27.0 & 35 & 43 \\
$\tau^3$-Banking & 11.6 & 12.89 & 14.69 & 5.03 & 7.22 & 8.63 & 13.66 & 9.54 \\
GDPval-AA v2 (Elo) & 815 & 946 & 967 & 424 & 546 & 657.4 & 751.6 & 987 \\
\rowcolor{black!5}
\multicolumn{9}{l}{\textbf{Reasoning}} \\
GPQA Diamond (no tools) & 69.19 & 72.41 & 72.16 & 77.71 & 80.93 & 73.36 & 80.43 & 85.35 \\
HLE (text-only, no tools) & 8.25 & 10.89 & 10.43 & 11.31 & 15.43 & 13.67 & 17.79 & 19.18 \\
\rowcolor{black!5}
\multicolumn{9}{l}{\textbf{Instruction Following}} \\
IFBench & 74.29 & 77.94 & 79 & 53.83 & 59 & 73.44 & 77.89 & 57.44 \\
\rowcolor{black!5}
\multicolumn{9}{l}{\textbf{Long Context}} \\
AA-LCR & 63 & 71 & 73 & 64.7 & 68.4 & 58.9 & 64.6 & 68.5 \\
GDP.pdf (text-only all-pass) & 3.4 & 6.2 & 6.0 & 2.8 & 4.6 & 4.00 & 3.60 & 4.40 \\
\rowcolor{black!5}
\multicolumn{9}{l}{\textbf{General Knowledge}} \\
AA-Omniscience (accuracy) & 10.03 & 10.08 & 10.03 & 15.02 & 18.05 & 17.20 & 21.13 & 19.95 \\
AA-Omniscience (non-hallucination) & 76.42 & 79.74 & 79.38 & 20.18 & 17.39 & 21.84 & 16.36 & 65.25 \\
\midrule
\textbf{Average (Equal weights)} & \textbf{35.17} & \textbf{39.67} & \textbf{40.60} & \textbf{28.53} & \textbf{32.21} & \textbf{32.59} & \textbf{36.73} & \textbf{42.30} \\
\bottomrule
\end{tabular}%
}
\caption[Post-trained evaluation results across inference loop counts]{
Evaluation results for \textbf{\ourmodel} across inference loop counts. All scores are percentages except GDPval-AA v2, which is reported in Elo. GDPval-AA v2 is normalized as $(\mathrm{Elo}-500)/20$ for unweighted average score over 12 benchmarks. Results are measured by NVIDIA using NeMo Gym / NeMo Evaluator, except
$\dagger$, which denotes a self-reported score. Terminal-Bench v2.1 uses the
Terminus-2 harness for all models. SciCode reports pooled subtask accuracy;
IFBench~\citep{pyatkin2025generalizing} reports mean constraint satisfaction under strict checking.
GDP.pdf uses text-only inputs and reports the all-pass rate averaged over
five repeats per task for \ourmodel.
}
\label{tab:posttraining_evals}
\end{table}

\paragraph{Coding and agentic capabilities.}
We evaluate coding and agentic capabilities using SWE-Bench Verified~\citep{jimenez2023swe,chowdhury2024swebenchverified} for software engineering agents; Terminal-Bench v2.1~\citep{merrill2026terminal} for terminal-based task execution; GDPval-AA v2~\citep{patwardhan2025gdpval} for economically valuable real-world workplace tasks; $\tau^3$-Banking~\citep{barres2025tau} for conversational tool use in banking; and SciCode~\citep{tian2024scicoderesearchcodingbenchmark} for scientific coding. We report SWE-Bench Verified as pass@1 (the fraction of issues resolved by the generated patch), Terminal-Bench v2.1 as pass@1, GDPval-AA v2 in Elo, $\tau^3$-Banking as task success rate, and SciCode as pooled subtask accuracy.

\paragraph{Reasoning and general knowledge.}
We evaluate expert-level knowledge and reasoning using GPQA Diamond~\citep{rein2023gpqa} and HLE~\citep{phan2025humanitysexam}, both without tools. We measure factual reliability using the judge accuracy and non-hallucination metrics of AA-Omniscience~\citep{jackson2025aaomniscience}. The latter is one minus the proportion of incorrect answers among all non-correct responses, including partially correct and abstained answers.

\paragraph{Long-context understanding.}
We evaluate long-context reasoning using AA-LCR~\citep{artificialanalysis2025lcr}, which measures information retrieval, synthesis, and reasoning over long inputs, and report judge accuracy (the fraction of responses judged correct). We also evaluate GDP.pdf~\citep{garre2026gdppdf}, which tests reasoning over long professional documents with task-specific grading criteria. As \ourmodel~is a text-based model, we use text-only document inputs and report the fraction of responses satisfying every grading criterion, averaged over five repeats per task.

\paragraph{Main results.} Table~\ref{tab:posttraining_evals} compares \ourmodel~at inference loop counts $k\in\{1,2,4\}$ with Qwen3.5-4B,\footnote{\url{https://huggingface.co/Qwen/Qwen3.5-4B}} Qwen3.5-9B,\footnote{\url{https://huggingface.co/Qwen/Qwen3.5-9B}} Gemma~4 12B,\footnote{\url{https://huggingface.co/google/gemma-4-12B-it}} Gemma~4 26B-A4B,\footnote{\url{https://huggingface.co/google/gemma-4-26B-A4B-it}} and Qwen3.6-35B-A3B\footnote{\url{https://huggingface.co/Qwen/Qwen3.6-35B-A3B}} across 12 benchmarks. \ourmodel~demonstrates improved agentic performance with deeper execution: GDPval-AA v2 increases from 815 to 967 Elo, SWE-Bench Verified from 50.3\% to 61.8\%, and Terminal-Bench v2.1 from 12.5\% to 22.2\% between one and four loops. However, gains from deeper execution are task-dependent: AA-Omniscience accuracy remains essentially unchanged across loop counts.
At four loops, \ourmodel~outperforms all evaluated baselines on $\tau^3$-Banking, IFBench, and AA-LCR, including models with 26B and 35B total parameters. It also exceeds Qwen3.5-9B on SciCode (35.11\% vs.\ 28.21\%), while approaching Qwen3.6-35B-A3B on GDPval-AA v2 (967 vs.\ 987 Elo). These results suggest that increasing recurrent depth enables \ourmodel~to compete with larger models on several coding, agentic, and long-context tasks without adding parameters.

\subsection{Token Efficiency}
\label{sec:posttraining_token_efficiency}

Additional loops not only improve accuracy on agentic tasks but also reduce the number of tokens the model generates to complete them. Table~\ref{tab:posttraining_token_efficiency} decomposes the output tokens per attempt into the number of turns per attempt and the output tokens per turn. Going from one to four loops, output tokens per attempt fall by 1.3--2.5$\times$, and both factors contribute, in proportions that depend on the task. On the long-horizon tasks, most of the savings come from completing the task in fewer turns: on GDPval-AA v2, turns per attempt drop from 116 to 49 while output tokens per turn change by only 5\%, and on Terminal-Bench v2.1, turns drop by 39\% and tokens per turn by 20\%. On SWE-Bench Verified, the two factors contribute comparably, with 17\% fewer turns and 11\% fewer tokens per turn. On $\tau^3$-Banking, where the conversation length is largely set by the simulated user, the number of turns changes by only 7\%, and most of the savings come from 18\% shorter turns. Thus, with additional loops, \ourmodel both takes fewer actions to complete a task and writes less per action. These savings offset much of the higher compute per token. Approximating generation compute as output tokens times layer applications per token (72, 112, and 192 at $k=1,2,4$), two loops cost only 1.03--1.25$\times$ the compute of one loop across the four tasks, yet recover most of the accuracy gain of four loops: 86\% of the GDPval-AA v2 Elo gain, 80\% on SWE-Bench Verified, 58\% on Terminal-Bench v2.1, and 42\% on $\tau^3$-Banking. Four loops, by contrast, cost 1.07--2.04$\times$ the compute of one loop. The second loop therefore gives the largest accuracy gain per unit of added compute, making $k=2$ a good default, with $k=4$ for tasks where accuracy matters most.

\begin{table}[t]
  \centering
  \small
  \setlength{\tabcolsep}{5pt}
  \renewcommand{\arraystretch}{1.3}
  \resizebox{0.95\textwidth}{!}{%
  \begin{tabular}{l|ccc|ccc|ccc}
    \toprule
    \multirow{2}{*}{\textbf{Task}}
      & \multicolumn{3}{c|}{\textbf{Turns per attempt}}
      & \multicolumn{3}{c|}{\textbf{Output tokens per turn}}
      & \multicolumn{3}{c}{\textbf{Output tokens per attempt}} \\
    \cmidrule(lr){2-4}\cmidrule(lr){5-7}\cmidrule(lr){8-10}
      & \textbf{$k=1$} & \textbf{$k=2$} & \textbf{$k=4$}
      & \textbf{$k=1$} & \textbf{$k=2$} & \textbf{$k=4$}
      & \textbf{$k=1$} & \textbf{$k=2$} & \textbf{$k=4$} \\
    \midrule
    GDPval-AA v2       & 116  & 76   & 49   & 1,832 & 1,836 & 1,740 & 212k  & 140k  & 85k   \\
    Terminal-Bench v2.1 & 134  & 100  & 82   & 815   & 773   & 655   & 109k  & 78k   & 54k   \\
    SWE-Bench Verified & 88   & 76   & 73   & 384   & 349   & 341   & 33.6k & 26.4k & 24.8k \\
    $\tau^3$-Banking   & 21.3 & 20.1 & 19.8 & 1,081 & 919   & 890   & 23.0k & 18.5k & 17.6k \\
    \bottomrule
  \end{tabular}%
  }
  \caption{Token efficiency of \ourmodel{} across inference loop counts $k$ on agentic benchmarks. Output tokens per attempt is the product of turns per attempt and output tokens per turn, and includes reasoning tokens across all agent turns. Going from $k=1$ to $k=4$ reduces output tokens per attempt by 1.3--2.5$\times$ (Figure~\ref{fig:intro_tokens}).}
  \label{tab:posttraining_token_efficiency}
\end{table}

\section{Quantization}
\label{sec:quantization}

\subsection{Overview}
\label{sec:quant:overview}

We study post-training quantization (PTQ) of \ourmodel, followed by quantization-aware
distillation (QAD)~\citep{qadnvfp4inference} where needed, to create efficient quantized versions of the trained BF16 checkpoint that preserve its accuracy.
We evaluate three weight-only quantization formats: MXFP8, NVFP4, and Q5\_K\_M (GGUF). For each format, we first apply PTQ to all weight matrices except the token embeddings and LM head, then apply QAD if the quantized model falls short of the target accuracy. All learned vectors and scalars remain in high precision.

\subsection{Post-Training Quantization (PTQ)}
We apply PTQ using a custom implementation of NVIDIA ModelOpt \citep{nvidia-modelopt}. For NVFP4, we use dynamic MSE scale selection: for each block, we select the scale that minimizes the MSE across all 126 positive E4M3 values. For the other formats we use standard maximum scaling.
We find that MXFP8 achieves accuracy comparable to the BF16 baseline, while NVFP4 and Q5\_K\_M exhibit accuracy gaps that we aim to reduce through QAD. 

\begin{table}[t]
\centering
\small
\setlength{\tabcolsep}{4pt}
\renewcommand{\arraystretch}{1.2}
\resizebox{\linewidth}{!}{%
\begin{tabular}{l|ccc|ccc|ccc|ccc}
\toprule
\textbf{Task} &
\multicolumn{3}{c|}{\textbf{BF16}} &
\multicolumn{3}{c|}{\textbf{MXFP8-PTQ}} &
\multicolumn{3}{c|}{\textbf{NVFP4-QAD}} &
\multicolumn{3}{c}{\textbf{Q5\_K\_M-QAD}} \\
\cmidrule(lr){2-4}
\cmidrule(lr){5-7}
\cmidrule(lr){8-10}
\cmidrule(lr){11-13}

& \textbf{1} & \textbf{2} & \textbf{4}
& \textbf{1} & \textbf{2} & \textbf{4}
& \textbf{1} & \textbf{2} & \textbf{4}
& \textbf{1} & \textbf{2} & \textbf{4} \\
\midrule

\rowcolor{black!5}
\multicolumn{13}{l}{\textbf{Coding \& Agentic}} \\
SciCode
& 27.30 & 34.94 & 35.11
& 27.78 &32.68 &33.98 &26.78 &33.81 & 34.85& 25.48& 33.51& 34.29\\
Terminal-Bench v2.1
& 12.5 & 18.1 & 22.2
&11.94 &17.56 &22.19 &15.03 &19.24 &20.08 &15.03 &19.24 &18.26 \\
$\tau^3$-Banking
& 11.60 & 12.89 & 14.69
& 12.78& 12.78& 14.43& 12.16& 15.88& 14.85&12.16 &12.78 & 15.05\\
GDPval-AA v2 (Elo)
& 815 & 946 & 967
& 856& 961& 985& 759& 929& 965& 804.5& 952.8& 985.2\\
\midrule

\rowcolor{black!5}
\multicolumn{13}{l}{\textbf{Reasoning}} \\
GPQA Diamond (no tools)
& 69.19 & 72.41 & 72.16
&69.51 &70.77 &71.91 &68.06 &72.16 &72.35 & 68.69& 72.54& 72.66\\
HLE (text-only, no tools)
& 8.25 & 10.89 & 10.43
&9.18 & 10.43& 10.98& 8.67& 9.92& 10.66& 9.5& 10.38& 11.91\\
\midrule

\rowcolor{black!5}
\multicolumn{13}{l}{\textbf{Long Context}} \\
AA-LCR
& 63.06 & 71.31 & 72.94
&65.69 & 71& 71.19& 57.5& 69.5&69.12 &62.88 &70.44 &70.12 \\
\midrule

\rowcolor{black!5}
\multicolumn{13}{l}{\textbf{General Knowledge}} \\
AA-Omniscience (accuracy)
& 10.03 & 10.08 & 10.03
&10.17 & 9.93& 9.7& 10.2& 10.4&10.2 &9.85 &9.7 &10.05 \\
AA-Omniscience (non-hallucination)
& 76.42 & 79.74 & 79.38
&77.44 &81.96 &82.24 &75.13 &79.95 &80.31 & 76.17& 80.29& 82.25\\
\bottomrule
\end{tabular}%
}
\caption[BF16 and quantized-model evaluation results]{
Evaluation results for \textbf{\ourmodel} in BF16 and its weight-only
quantized variants: MXFP8-PTQ, NVFP4-QAD, and Q5\_K\_M-QAD,
at one, two, and four inference loops.
We report a subset of the benchmarks in Table~\ref{tab:posttraining_evals}, using the same metrics.
}
\label{tab:bf16_ptq_qad_evals}
\end{table}

\subsection{Quantization-Aware Distillation (QAD)}

\paragraph{Data blend.} We reuse the original SFT mixture from Section~\ref{sec:sft} and create a weighted subsample of 10B raw input tokens, with $\sim$3B stored assistant mask tokens. 

\paragraph{QAD setup.} The frozen teacher is the original BF16 checkpoint corresponding to the
student. The student is the PTQ version and applies
the selected weight quantizer in every forward pass. Both models receive
the same SFT token sequence and the same recurrent-loop count.

With teacher and student distributions $p_T$ and $p_S$, respectively,
we minimize full-vocabulary forward KL at temperature one:
\begin{equation}
 \mathcal L_{\mathrm{KD}}=
 \frac{1}{\sum_t m_t}\sum_t m_t\sum_{v\in\mathcal V}
 p_T(v\mid x_{\leq t},k)
 \log\frac{p_T(v\mid x_{\leq t},k)}
          {p_S(v\mid x_{\leq t},k)}.
\end{equation}
Here $m_t$ marks valid assistant next-token targets. 
Prompt, system, user,
document, padding, and invalid conversation-boundary targets do not
contribute to the objective, although their context still affects assistant
predictions. 

During QAD, we unfreeze all parameters, including embeddings, the LM head and vectors. 
For NVFP4, global and block scales are recomputed from the current weights
on each quantized forward. 
We follow the original SFT-style training regime including the loop sampler, truncated backpropagation through loops (TBPTL) with depth one and matrix-normalization after each optimizer update.
We use AdamW with $\beta_1=0.9$, $\beta_2=0.95$, and a cosine schedule with linear warmup and a peak learning rate of $10^{-5}$. We use a global batch
of 128 packed sequences, a maximum context of 256K tokens, eight-way context parallelism, and 16-way data parallelism. 
We run QAD on 128 NVIDIA GB300 GPUs for 12 hours for each of the candidates.

\subsection{Results}
\label{sec:quant:results}

\paragraph{Quantization results.} Table~\ref{tab:bf16_ptq_qad_evals}
compares the BF16 baseline with MXFP8-PTQ, NVFP4-QAD, and Q5\_K\_M-QAD
at inference loop counts $k\in\{1,2,4\}$. MXFP8-PTQ remains broadly
comparable to BF16 without additional distillation. The QAD variants also
retain competitive performance across the benchmark suite, largely
preserving the BF16 model's capabilities at lower weight precision, with
some task-dependent accuracy gaps.

\paragraph{Performance.} Table~\ref{tab:quantization_speedup} reports measured inference speedups for \ourmodel on NVIDIA DGX Spark for MXFP8 and NVFP4 datatypes relative to the BF16 baseline. Quantization accelerates inference by up to 2.44$\times$ over the BF16 baseline. 

\begin{table}[t]
\centering
\small
\setlength{\tabcolsep}{8pt}
\renewcommand{\arraystretch}{1.2}
\resizebox{\linewidth}{!}{%
\begin{tabular}{ccccc}
\toprule
\shortstack{\textbf{Input sequence}\\\textbf{length (ISL, tokens)}} &
\shortstack{\textbf{Output sequence}\\\textbf{length (OSL, tokens)}} &
\shortstack{\textbf{Batch}\\\textbf{size}} &
\shortstack{\textbf{MXFP8 (W8A16)}\\\textbf{speedup vs.\ BF16}} &
\shortstack{\textbf{NVFP4 (W4A16)}\\\textbf{speedup vs.\ BF16}} \\
\midrule
2560 & 300  & 1 & $1.69\times$ & $2.17\times$ \\
128  & 1024 & 1 & $1.84\times$ & $2.44\times$ \\
\bottomrule
\end{tabular}%
}
\caption{Inference speedup relative to BF16 for MXFP8 and NVFP4
weight-only quantization across two sequence-length configurations.}
\label{tab:quantization_speedup}
\end{table}

\section{Inference Optimization}
\label{sec:inference_optimization}

We explore two ways to improve inference efficiency in \ourmodel{}. Last-loop KV caching makes attention KV memory independent of loop count, while self-speculative decoding uses shallow drafting and deeper verification to accelerate generation. We evaluate these optimizations separately, using per-loop KV caches for self-speculative decoding.

\subsection{Last-Loop KV Caching}
\label{sec:posttraining_last_loop_kv}

With standard per-loop caching, each recurrent attention layer maintains a separate KV cache for every loop: a loop-$i$ query attends to earlier tokens' loop-$i$ keys and values. The recurrent block's KV memory therefore grows linearly with the loop count $k$, while the input and output blocks each retain one cache.
Last-loop KV caching~\citep{ladhak2026elr} instead maintains a single cache per recurrent attention layer, overwriting its entries at each loop. After each forward step, the cache retains the final-loop keys and values. Queries attend to final-loop entries for previously cached tokens and current-loop entries for tokens processed in the same step. This keeps attention KV memory at its one-loop footprint for every $k$, without retraining, but exposes earlier-loop queries to final-loop keys and values rather than the same-loop entries used during training.

We compare both cache policies on the final post-trained checkpoint at $k=4$, with $k=1$ as a reference, where the policies coincide (Table~\ref{tab:posttraining_last_loop_kv}). Serving uses prefix caching and chunked prefill. Last-loop caching retains gains from additional loops, with score changes of $-2.12$ to $+0.42$ points relative to per-loop caching at $k=4$. Mean output lengths change little, from 1.0\% fewer tokens on GPQA Diamond to 3.4\% more on Terminal-Bench v2.1. Last-loop caching therefore makes attention KV memory independent of loop count while retaining gains from deeper execution, with little change in generation length.

\begin{table}[t!]
  \centering
  \setlength{\tabcolsep}{6pt}
  \renewcommand{\arraystretch}{1.15}
  \resizebox{0.95\textwidth}{!}{%
  \begin{tabular}{lr|ccccc}
    \toprule
    \multirow{2}{*}{\textbf{Task}} & \multirow{2}{*}{\textbf{Metric}}
      & \multicolumn{2}{c}{\textbf{Per-loop}}
      & \textbf{Last-loop}
      & \multirow{2}{*}{\shortstack{\textbf{$\Delta$}\\[1pt]\textbf{Score}}}
      & \multirow{2}{*}{\shortstack{\textbf{Output}\\[1pt]\textbf{Tokens}}} \\
    \cmidrule(lr){3-4}\cmidrule(lr){5-5}
      & & \textbf{$k=1$} & \textbf{$k=4$} & \textbf{$k=4$} & & \\
    \midrule
    $\tau^3$-Banking & \textit{task success rate, $n=8$}  & 11.60 & 14.69 & 14.30 & $-0.39$ & $1.00\times$ \\
    Terminal-Bench v2.1 & \textit{pass@1, $n=8$}  & 12.50 & 22.19 & 20.22 & $-1.97$ & $1.03\times$ \\
    \midrule
    GPQA Diamond & \textit{accuracy, no tools, $n=8$}  & 69.19 & 72.16 & 72.10 & $-0.06$ & $0.99\times$ \\
    \midrule
    SciCode & \textit{subtask accuracy, $n=8$}  & 27.30 & 35.11 & 35.37 & $+0.26$ & $1.01\times$ \\
    \midrule
    AA-Omniscience (accuracy) & \textit{judge accuracy, $n=10$}  & 10.03 & 10.03 & 10.45 & $+0.42$ & $1.03\times$ \\
    AA-Omniscience (non-halluc.) & \textit{$1-\text{hallucination rate}$, $n=10$}  & 76.42 & 79.38 & 79.25 & $-0.13$ & $1.03\times$ \\
    \midrule
    AA-LCR & \textit{judge accuracy, $n=16$} & 63.06 & 72.94 & 70.81 & $-2.12$ & $1.02\times$ \\
    \bottomrule
  \end{tabular}%
  }
  \caption{\ourmodel{} with per-loop and last-loop KV caches. Scores are percentages and $n$ is the number of repeats per task. $\Delta$ Score is last-loop minus per-loop at $k=4$, computed before rounding. Output Tokens is the ratio of mean output tokens per rollout, last-loop to per-loop, at $k=4$, including reasoning and all agent turns or sub-steps and excluding judge and user-simulator tokens.}
  \label{tab:posttraining_last_loop_kv}
\end{table}

\subsection{Self-Speculative Decoding}
\label{sec:speculative_decoding}

\ourmodel{} uses its variable execution depth for self-speculative decoding, with the same checkpoint drafting at $k=1$ and verifying at $k=4$. The drafter proposes $\gamma=4$ tokens autoregressively, which the verifier evaluates in a batched forward pass. Rejection sampling uses both models' probabilities to accept proposals and correct rejections, preserving the verifier's sampling distribution in exact arithmetic without a separate draft model.

With per-loop caching, drafting and verification share the KV caches of the input block and first recurrent loop. We also reuse the corresponding draft activations during verification, avoiding redundant computation for token positions already processed by the drafter. Drafting and verification execute sequentially, so the acceleration comes from accepting multiple tokens per verification step and reusing computation.

\begin{figure}[t]
  \centering
  \includegraphics[width=0.5\linewidth]{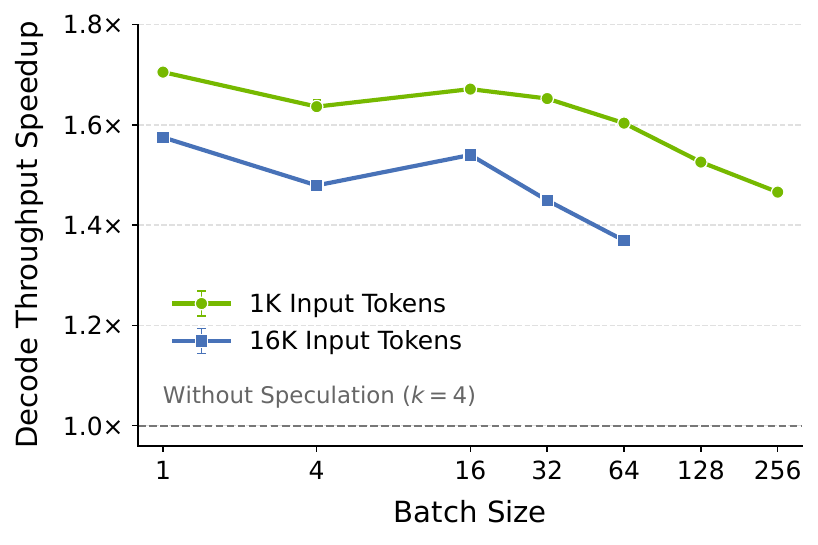}
  \caption{Self-speculative decode throughput relative to autoregressive $k=4$, using $k=1$ drafting, $k=4$ verification, and four draft tokens. Curves compare KV reuse with and without activation reuse. Throughput is measured over output positions 128--896 of 1,024-token generations. Points show geometric means across three repeat blocks and error bars show block ranges.}
  \label{fig:spec_batch_speedup}
\end{figure}

We first measure steady-state decode throughput across batch sizes and prompt lengths (Figure~\ref{fig:spec_batch_speedup}). With KV and activation reuse, self-speculation achieves $1.37$--$1.71\times$ the throughput of autoregressive decoding at $k=4$ across all measured configurations. Gains generally decrease at larger batch sizes but persist through batch size 256 for shorter prompts and batch size 64 for 16K-token prompts.

Self-speculative decoding reduces mean prefill-plus-decode request duration relative to autoregressive decoding at $k=4$, using the same checkpoint, benchmark sampling settings, and serving configuration (Table~\ref{tab:posttraining_self_spec}). Across six benchmarks, self-speculative decoding achieves $1.11$--$1.53\times$ speedups in mean prefill-plus-decode request duration, with gains depending on the workload and serving conditions.
\begin{table}[h!]
  \centering
  \small
  \setlength{\tabcolsep}{10pt}
  \renewcommand{\arraystretch}{1.3}
  \begin{tabular}{lc}
    \toprule
    \textbf{Task} & \textbf{Prefill + Decode Speedup} \\
    \midrule
    $\tau^3$-Banking  & $1.11\times$ \\
    Terminal-Bench v2.1 & $1.17\times$ \\
    GPQA Diamond      & $1.34\times$ \\
    SciCode           & $1.35\times$ \\
    AA-Omniscience     & $1.53\times$ \\
    AA-LCR     & $1.44\times$ \\
    \bottomrule
  \end{tabular}
  \caption{Self-speculative decoding speedups for \ourmodel{}, using $k=1$ drafting, $k=4$ verification, $\gamma=4$ draft tokens, and shared KV caches and activations. The baseline uses autoregressive decoding at $k=4$. Speedup is the ratio of baseline to speculative mean model-request duration, including prefill and decode and excluding initialization, initial queueing, tool execution, and evaluator processing.}
  \label{tab:posttraining_self_spec}
\end{table}

\section{Conclusion}
\label{sec:conclusion}
We present \ourmodel, a 4-billion-parameter looped language model with elastic depth, in which a single checkpoint serves different inference compute budgets by varying the number of times its shared recurrent block is applied. \ourmodel uses a normalized Transformer backbone with interleaved sliding-window and full-attention layers. It was pretrained on 6 trillion tokens, the first 4.5 trillion in NVFP4, extended to a 256K-token context, and post-trained with SFT and RLVR using a difficulty-aware loop curriculum. Additional loops improve accuracy, with the largest gains on coding, agentic, and long-context benchmarks; with two or more loops, \ourmodel outperforms larger models such as Qwen3.5-9B, Gemma~4 12B, and Gemma~4 26B-A4B on average across 12 benchmarks. Looping also makes \ourmodel more token-efficient: four loops generate 1.3--2.5$\times$ fewer tokens than one loop on agentic benchmarks, and the second loop gives the largest accuracy gain per unit of added compute. Last-loop KV caching keeps the KV-cache footprint equal to that of a non-looped model at any loop count, and MXFP8, NVFP4, and GGUF quantized checkpoints further reduce deployment cost. We open-source the BF16 and quantized checkpoints of \ourmodel.

\bibliography{references}
\bibliographystyle{references}

\end{document}